\documentclass{article}
\usepackage{ijcai24}

\usepackage{times}
\usepackage{soul}
\usepackage{url}
\usepackage[hidelinks]{hyperref}
\usepackage[utf8]{inputenc}
\usepackage[small]{caption}
\usepackage{graphicx}
\usepackage{amsmath}
\usepackage{amsthm}
\usepackage{booktabs}
\usepackage{multirow}
\usepackage[linesnumbered,ruled]{algorithm2e}
\usepackage[switch]{lineno}

\def\eg{\emph{e.g.}} 
\def\ie{\emph{i.e.}}

\def\etal{\emph{et al.}}

\title{XPose: eXplainable Human Pose Estimation}

\author{
Luyu Qiu$^1$\and
Jianing Li$^2$\and
Lei Wen$^1$\and
Chi Su$^2$\and
Fei Hao$^1$\and
Chen Jason Zhang$^{2}$\And
Lei Chen$^{1,2}$\\
\affiliations
$^1$Hong Kong University of Science and Technology\\
$^2$Hong Kong Polytechnic University\\
\emails
lqiuag@connect.ust.hk,
tensor.li@polyu.edu.hk,
lwen@connect.ust.hk,
{chisu, ffaye.hao, jason-c.zhang}@polyu.edu.hk,
leichen@cse.ust.hk
}

\begin{document}

\maketitle

\begin{abstract}
Current approaches in pose estimation primarily concentrate on enhancing model architectures, often overlooking the importance of comprehensively understanding the rationale behind model decisions.
In this paper, we propose XPose, a novel framework that incorporates Explainable AI (XAI) principles into pose estimation. This integration aims to elucidate the individual contribution of each keypoint to final prediction, thereby elevating the model's transparency and interpretability.
Conventional XAI techniques have predominantly addressed tasks with single-target tasks like classification. Additionally, the application of Shapley value, a common measure in XAI, to pose estimation has been hindered by prohibitive computational demands.

To address these challenges, this work introduces an innovative concept called Group Shapley Value (GSV). This approach strategically organizes keypoints into clusters based on their interdependencies. Within these clusters, GSV meticulously calculates Shapley value for keypoints, while for inter-cluster keypoints, it opts for a more holistic group-level valuation. This dual-level computation framework meticulously assesses keypoint contributions to the final outcome, optimizing computational efficiency.
Building on the insights into keypoint interactions, we devise a novel data augmentation technique known as Group-based Keypoint Removal (GKR). This method ingeniously removes individual keypoints during training phases, deliberately preserving those with strong mutual connections, thereby refining the model's predictive prowess for non-visible keypoints. The empirical validation of GKR across a spectrum of standard approaches attests to its efficacy. GKR's success demonstrates how using Explainable AI (XAI) can directly enhance pose estimation models.

\end{abstract}


\section{Introduction}
Recent years witness the rapid develop of deep learning technology on various of field include computer vision~\cite{krizhevsky2012imagenet,he2016deep} and neural language process~\cite{vaswani2017attention,devlin2018bert}.
However, the increasing complexity makes model challenging to design through manual intuition. Therefore, it's crucial to understand model's decision-making process.

The emergence of eXplainable AI (XAI) has addressed the opacity and lack of interpretability associated with advanced machine learning models~\cite{ijcai2020p726}, enabling researchers and practitioners to comprehend the rationale behind their predictions \cite{Visualizing2014}.
Existing XAI methods have primarily concentrated on single-target tasks like classification~\cite{petsiuk2018rise} and detection \cite{sarp2023xai}, offering valuable insights into the decision-making processes of these models. However, as AI tasks become more intricate, the inherent correlations in multi-target tasks are often neglected, limiting the applicability of existing XAI techniques.

\begin{figure}
\centering
\includegraphics[width=1\linewidth]{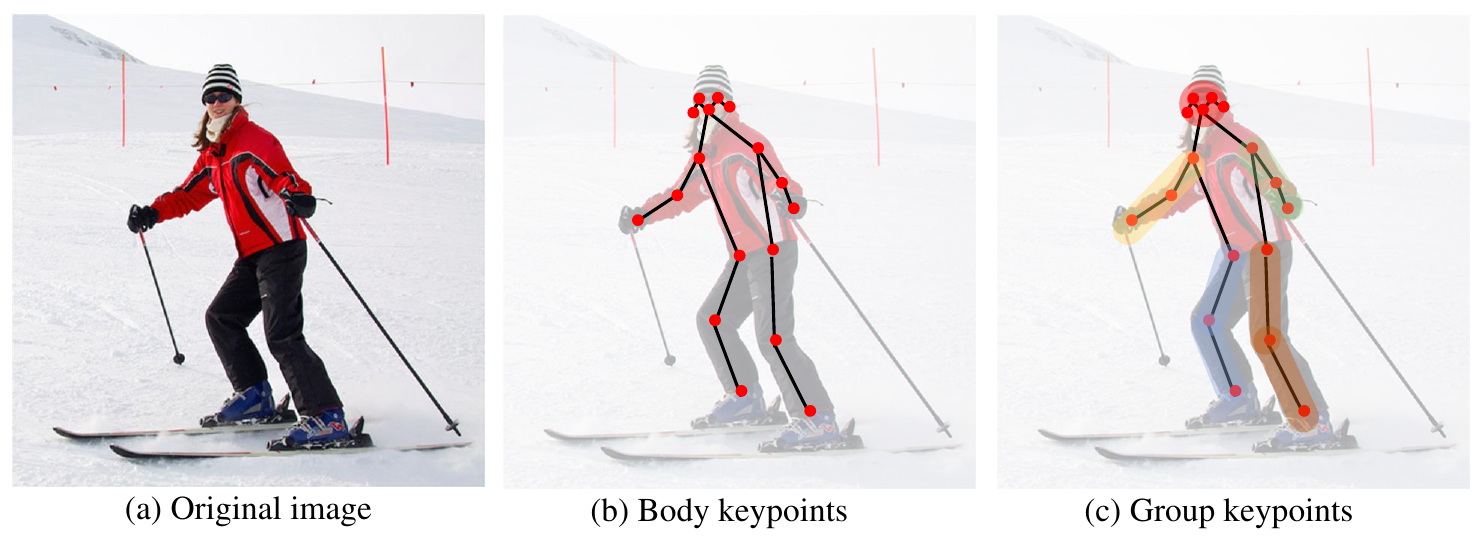}\\
\caption{Illustrations of (a) human image, and corresponding (b) human body keypoints and (c) groupped keypoints.}
\label{fig:example}
\end{figure}

Multi-Person Pose Estimation (MPPE) aims to identify and localize human body keypoints. Fig.\ref{fig:example}(a-b) illustrates the MPPE task.
As a fundamental yet challenging task in human behavior understanding, pose estimation has attracted increasing attention in recent years.
In MPPE task, the prediction of individual body keypoint will be influenced by each other keypoints. However, existing methods regard body keypoints as separate ones, and mainly concentrate on design superior models~\cite{wei2016convolutional,he2017mask,newell2017associative,cheng2020higherhrnet} or unbiased keypoint formulation~\cite{zhou2019objects,zhang2020distribution}, overlooking the correlation between keypoints, which could be beneficial for keypoint predictions.
On the other hand, current XAI methods are mainly designed for single-target tasks, may fall short in capturing the intricate relationships. Hence, these methods are challenging to directly apply for MPPE models.

This paper focuses on revealing the decision-making process of MPPE models through elucidating the individual contribution of each keypoint to final predictions, offering insights for future model design.
In MPPE task, the black-box model aims to predicts biological body keypoints of each person presents in input image.
The shapley value~\cite{shapley1951notes} is a common measure in XAI which used to quantify the contribution of each feature by calculating the average marginal impact across all possible feature combinations.
However, the abundance of keypoints poses a challenge as computing shapley value for each individual keypoint requires evaluating all possible combinations of keypoints, which will cause unaffordable computation cost. 
To handle this difficulty, we design a Group Shapley Value (GSV), to compute the contributation of each keypoint in a coarse-to-fine manner.
As shown in Fig.~\ref{fig:example}(b), a body keypoint only biologically connects to a limited number of keypoints, hence we compute fine-grained shapley value for only closely related keypoints, through which avoid redundant computations. While for loosely related keypoints, GSV calculates coarse-grained group-level shapley value to reduce the computation cost. Both fine-grained and coarse-grained constitute to overall contribution of each keypoint to final contribution. 
To group keypoints, we first estimate the interdependencies of each keypoint pair.
Following RISE~\cite{petsiuk2018rise}, we assess the 
interdependencies of each two keypoints through removing pixels around keypoints in the image.
Since the prior information from physiological structure of human body is also important, we introduce connectivity of keypoints when evaluating interdependencies. Fig.~\ref{fig:example}(c) illustrates the group result.

Building on the insights into keypoint interactions, we design Group-based Kepoint Removal (GKR) data augmentation to enhence model's ability of inferring invisible keypoint. 
During training, GKR randomly remove individual keypoint while keep its related keypoints unchanged, the loss of single keypoint compels model to infer through other closely related ones. Consequently, the model learns to better deduce invisible keypoints during inference. 

The contribution of this work can be summarized into
following aspects:
1) We propose XPose, a XAI method to reveal the decision make process of MPPE task. To best of our knowledge, this is an original work introducing XAI into MPPE.
2) A group shapley value is designed to reduce the computation complexity in shapley value, makes it better fit pose estimation task.
3) Based on interpretation of XPose, we further design a Group-based Kepoint Removal to enhence the model's ability of inferring invisible keypoints, and significantly boosts the performance of several typical methods.

\section{Related work}

This work is closely related to eXplainable AI(XAI) and human pose estimation. This section briefly summarizes those two categories of works.

\subsection{eXplainable AI}

Gradient-based XAI methods offer detailed insights into model decision-making by requiring the model to be a white-box and its processes differentiable, which limits their real-world applicability. In contrast, perturbation-based XAI methods, which do not assume the model to be transparent or differentiable, have broader applications due to their versatile techniques like addition, removal, and replacement, with removal being the most utilized \cite{mypaper}.

A recent survey \cite{removal_based_XAI_survey} provided a comprehensive overview of removal-based XAI methods, proposing a framework that dissects these methods through feature removal methods, model behavior, and summary technique. \textbf{Feature removal} techniques are characterized by diverse approaches to substituting altered segments, each with distinct methodologies for recalculating or replacing feature values. 
Simple methods such as zero \cite{petsiuk2018rise} or default value \cite{lime} directly modify feature values with zero or a default value. In contrast, missingness during training \cite{yoon2018invase} conditions the model to interpret zeros or alternative replacement values as indicators of missing data, rather than features naturally equating to zero. Techniques like blurring with a Gaussian kernel \cite{DBLP:conf/iccv/FongPV19} or extending pixel values from neighboring pixels \cite{zhou2015object} offer more nuanced alterations. Generative model \cite{chang2019explaining} approaches calculate the fill values through generative techniques whereas separate models \cite{SPVIM} replace the single model with separate models for every feature.

\textbf{Model behavior} techniques can be categorized into three types. Prediction methods evaluate the impact of excluding specific features on the individual prediction \cite{Visualizing2014}. Prediction loss methods measure the prediction loss \cite{TreeSHAP} or prediction mean loss \cite{yoon2018invase} calculated by true labels and prediction labels. Dataset loss methods evaluate the overall performance alteration of a model when every features removed systematically \cite{SPVIM}.

\textbf{Summary techniques} encompass several approaches. Shapley value
quantifies an individual feature's contribution by considering all possible permutations \cite{shapleyValue}. This paper also involves the concept of Shapley value. Techniques for removing \cite{Visualizing2014} or including \cite{10.5555/944919.944968} individual features assess the impact of these actions on model performance. Additive models employ regularized additive models to analyze the dataset comprised of perturbed examples \cite{lime}. High-value subset and low-value subset formulate the summarization as an optimization problem. The former \cite{zhou2015object} seeks a feature subset that maximizes the set function's value, whereas the latter \cite{DBLP:conf/iccv/FongPV19} selects features whose exclusion minimizes the set function's value.

\begin{figure*}
\centering
\includegraphics[width=0.9\linewidth]{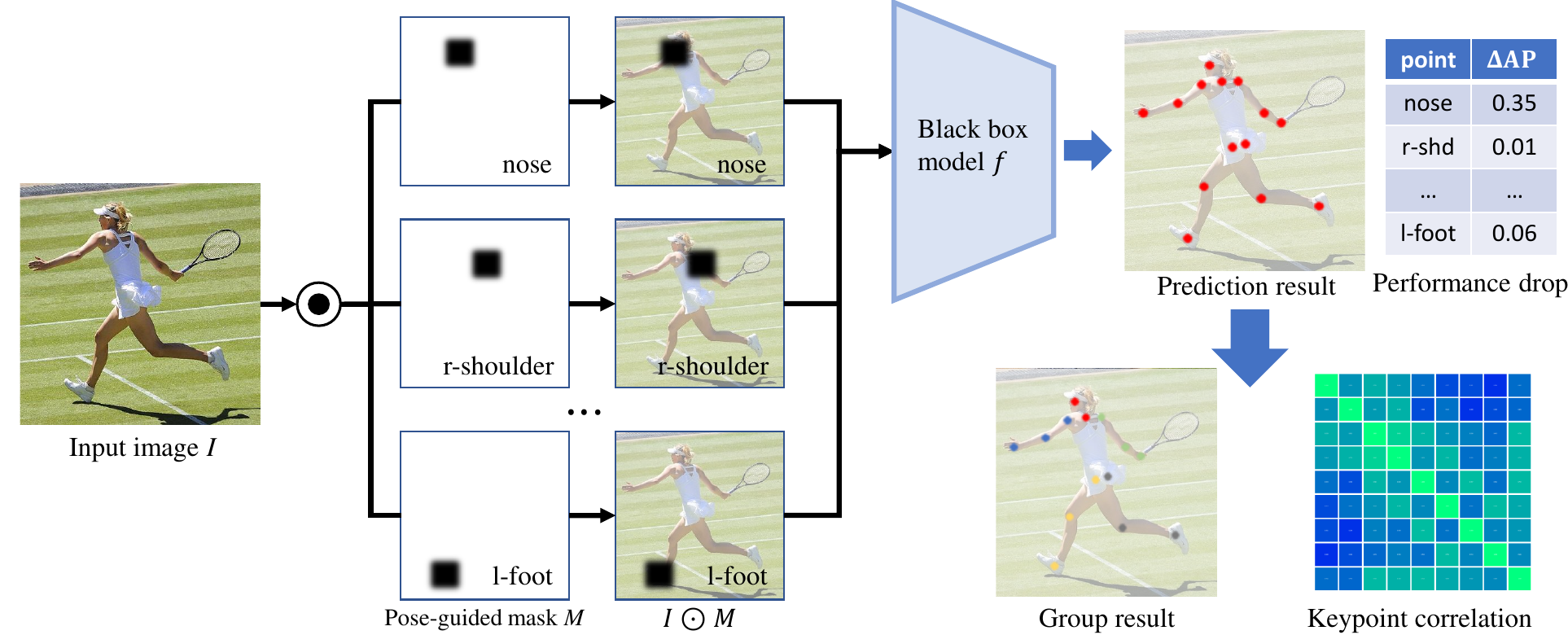}\\
\caption{Overview of the proposed XPose for explainable pose estimation. XPose takes masked person image as input and evaluate the influence of each keypoint on all other keypoints.}
\label{fig:framework}
\end{figure*}

\subsection{Pose estimation}

As a fundamental yet challenging task in human behavior understanding, pose estimation has attracted increasing attention in recent years. Based on the keypoint formulation, existing works can be summarized into two categories, \ie, heatmap-based methods and regression based methods.

\textbf{Heatmap based pose estimation} predicts a probability for each person and locate the keypoint through taking the maximum of heatmap. Based on the order of keypoints localization and grouping, heatmap based pose estimation works can be summarised into top-down and bottom-up methods
Top-down methods~\cite{he2017mask,fang2017rmpe,sun2018integral,sun2019deep} first detect and crop bounding box for each person in image, and resize cropped images to the same size before performing single person pose estimation.
He \etal~\cite{he2017mask} add a pose estimation branch based on Faster R-CNN~\cite{ren2015faster} and reuse the features for multi-task learning.
Fang \etal~\cite{fang2017rmpe} propose a symmetric spatial transformer network to refine the inaccurate bounding boxes.
The top-down models achieve promising performance on various MPPE datasets. However, the top-down paradigms need to crop, resize and estimate pose for every person instance individually, which significantly increases the computation and inference time.

In contrast, bottom-up methods~\cite{cao2017realtime,newell2017associative,cheng2020higherhrnet} first localize all keypoints in the image, then group them into different persons.
OpenPose~\cite{cao2017realtime} introduces a two-branch network to predict the heatmap for body joints and affinity field, respectively. The grouping is implemented through integrating two keypoints on affinity fields and grouping the keypoint pairs with the largest integral.
Cheng \etal~\cite{cheng2020higherhrnet} design a HigherHRNet to predict high resolution heatmap for keypoint estimation to solve the scale variation challenge in bottom-up methods.
Bottom-up method does not rely on extra human detectors, thus is more robust to occlusions. The grouping process could be complicated and time-consuming because it is hard to be processed in parallel.

\textbf{Regression based pose estimation}~\cite{zhou2019objects,nie2019single,wei2020point} formulates MPPE as a regression task, where human keypoints are defined as person center point and a set of offsets.
Zhou \etal~\cite{zhou2019objects} first regress the offset between body keypoints and person center, then match the regression result to the closest joint detected from heatmap to refine the result.
Instead of directly regressing from the center point, Wei \etal~\cite{wei2020point} define a set of anchor points and regress the keypoints from anchor points to avoid long-range regression.
Li \etal~\cite{li2023polarpose} try the factorize the 2D regression into angle and length in polar coordinate to simplify the regression through classification.
Compared with the two-stage heatmap based methods, regression based methods show better efficiency.

However, all above methods treat body keypoints as isolated entity, overlooking the inherent correlation, which is crucial for MPPE task. 
In this paper, we focus on XAI and reveal the intrinsic connections between body keypoints in decision making process. We believe this will provide guidance for the design of future pose estimation methods.

\section{Method}
\subsection{Formulation}
Given an image $I$, Multi-Person Pose Estimation (MPPE) aims to estimate the pose for each person via locating coordinates of body keypoints $\mathcal{P}$, $\mathcal{P}$ can be represented as,
\begin{equation}
\begin{aligned}
\mathcal{P}=\{k_1, k_2,...k_n\},
\end{aligned}
\label{equ:joint1}
\end{equation}
where $n$ is the number of keypoints. $k_i=(x_i,y_i,v_i)$ is the $i$-th keypoint, where $(x_i,y_i)$ and $v_i$ are coordinate and visibility of $i$-th keypoint, respectively.

For each keypoint $k_i$, XPose aims to evaluate the influence and significance of other keypoints on $k_i$. 
\begin{equation}
\begin{aligned}
X(k_i) = \{ \sigma(k_i, k_1), \sigma(k_i, k_2), ..., \sigma(k_i, k_n) \},
\end{aligned}
\label{equ:joint2}
\end{equation}
where $\sigma(k_i, k_j)$ denotes the contributation of keypoint $k_j$ when predicting keypoint $k_j$, and the contribution $\sigma$ satisfy 
 
\begin{equation}
\begin{aligned}
\sigma(k_i, k_j)& \geq 0\\
\sum_{j=1}^n(\sigma(k_i, \cdot))&=1.
\end{aligned}
\label{equ:joint3}
\end{equation}

The shapley value is a common measure in XAI to measure the contribution of each input feature. However, shapley value need to compute all combination of input features, which is unaffordable for high input dimensions.
To simplify the computation, we propose to compute shapley value in a coarse to fine manner.
We first cluster the keypoints into several groups based on their interdependencies, where 1) keypoints inside same group are closely-related with each other; 2) keypoints in different groups are loosely-related. Based on the group, we compute fine-grained shapley value in each group and group-level coarse-grained shapley value. 

Beside, with the guidance of keypoint interaction from XPose, we design a Group-based Keypoint Removal data augmentation to enhance model's inferential capability for invisible keypoints. Fig.~\ref{fig:framework} illustrates the overall framework of XPose. The following parts proceed to introduce details of XPose, including grouping process and group shapley value.

\subsection{Grouping process}
~\label{sec:group}
To cluster keypoints, we evaluate interdependencies between each pair of keypoints
One way to measure the interdependencies of a keypoint is to perturb it and observe how much this affects the model's decision of other keypoint. RISE~\cite{petsiuk2018rise} try to explore the importance of image areas through applying thousands of randomly generated masks on image and observe the output of black-box models.

Randomly perturb input image and observe output is intuitionistic but ineffective, especially for MPPE task, as majority of regions in image are unrelated background.
Considering keypoint's location is highly related with specific local region in image, we guide perturbation using the coordinates of keypoints. Specially, to evaluate the influence of keypoint $k_j$ to keypoint $k_i$, we randomly generate $m$ masks with random aspect ratio and area, and average the performance of keypoint $k_i$ as the perturbation $\Delta p(k_i,k_j)$. 
We normalize $\Delta p(k_i,k_j)$ as perturbation influence of each keypoint pair,

\begin{equation}
\begin{aligned}
\operatorname{PI}(k_1, k_2) = \frac{1}{2}({\frac{\Delta p(k_i,k_j)}{\sum_{a=1}^n\Delta p(k_i,k_a)} +  \frac{\Delta p(k_j,k_i)}{\sum_{a=1}^n\Delta p(k_j,k_a)}})
\end{aligned}
\label{equ:relationship1}
\end{equation}
where $\Delta p(k_i,k_j)$ denotes the performance drop of keypoint $k_i$ when perturbing $k_j$.
A higher $PI$ denotes the two keypoints are highly related.

Besides, we consider the structural information of the human body. Physiologically connected keypoints exhibit stronger interdependencies, whereas is weaker, \eg, the elbow keypoint has closer connection with shoulder and wrist keypoints, but weaker connection with knees.
We then incorporate keypoint connectivity into interdependencies.
For keypoint pair $(k_1, k_2)$, we define the keypoint connectivity as,
\begin{equation}
\begin{aligned}
\operatorname{KC}(k_i, k_j) = \frac{1}{2}({\frac{conn(k_i,k_j)}{\sum_{a=1}^nconn(k_i,k_a)} +  \frac{conn(k_j,k_i)}{\sum_{a=1}^nconn(k_j,k_a)}})
\end{aligned}
\label{equ:relationship2}
\end{equation}
where $conn(k_i,k_j)$ is connectivity of two keypoints, $conn=1$ if 2 keypoints are biologically connected else $0$. Similarly, we normalize the connectivity to keypoint connectivity.

The interdependencies $s_{inter}$ of keypoint pair jointly considers perturbation influence and keypoint connectivity,
\begin{equation}
\begin{aligned}
s_{inter}(k_1, k_2) = \operatorname{PI} + \operatorname{KC}
\end{aligned}
\label{equ:relationship3}
\end{equation}
Based on $s_{inter}$, we group kypoints and the grouping results are summarised in Table~\ref{table:group}. 
In Table~\ref{table:group}, keypoint are clustered into 5 groups, which aligns well with human body structure, \eg, head, arms and legs. The grouped keypoints are subsequently used to compute group shapley value.

\begin{table}
\caption{The group result of keypoints.}
\setlength{\tabcolsep}{25 pt}
\small
\begin{tabular}{l|c}
\toprule
Group id  &keypoint name   \\
\midrule
Group 1 &l-eye, r-eye, nose, l-ear, r-ear \\
Group 2 &l-shoulder, l-elbow, l-wrist \\
Group 3 &r-shoulder, r-elbow, r-wrist \\
Group 4 &l-hip, l-knee, l-foot \\
Group 5 &r-hip, r-knee, r-knee \\
\bottomrule
\end{tabular}
\label{table:group}
\end{table}

\subsection{Group Shapley Value}

Directly compute shapley value on all keypoints will suffer unaffordable computation cost. Considering each keypoint is related with only by a subset of keypoints, we propose Group Shapley Value (GSV) to compute shapley value in a coarse-to-fine manner.

GSV first cluster all keypoints $\mathcal{P}$ into $g$ groups based on interdependencies $s_{inter}$ in Sec.~\ref{sec:group},
\begin{equation}
\begin{aligned}
\operatorname{Cluster}(\mathcal{P}) = \{\mathcal{G}_1, \mathcal{G}_2,...,\mathcal{G}_g\}.
\end{aligned}
\label{equ:relationship4}
\end{equation}
GSV then computes fine-grained shapley value for keypoints within same group $\mathcal{G}$, in which keypoints are highly interdependent. And course-grained group-level shapley value for keypoints cross groups which are weakly interdependent.

For keypoint $k_i$, we evaluate the contribution of keypoint $k_j$ to predict $k_i$ through computing fine-grained shapley value within group, 
\begin{equation}
\begin{aligned}
 \phi_{ij} = \sum_{g\in\mathcal{G}} \omega(|g|)[v_i(g)-v_i(g\setminus \{j\})]
\end{aligned}
\label{equ:shapley1}
\end{equation}
where $g$ is a keypoint subset of group $\mathcal{G}$, $v_i(g)$ denotes the performance of keypoint $k_i$ with the keypoints in set $g$ visible, and $v_i(s\setminus \{j\})$ denotes the performance with keypoint $k_j$ perturbed, $\omega(|g|)$ is the weight of subset $g$.

And the group level shapley value $\varphi_{ij}$ treat each group as individual input, and compute the contribution of other groups to group $\mathcal{G}_i$,
\begin{equation}
\begin{aligned}
 \varphi_{ij} = \sum_{\mathcal{G}\in\operatorname{Cluster(\mathcal{P})}} \omega(|g|)[v_i(g)-v_i(g\setminus \{j\})],
\end{aligned}
\label{equ:shapley2}
\end{equation}
where $v_i(\mathcal{G}$ denotes the average performance of keypoints in group $\mathcal{G}_i$ with subset $g$ visible.
Both keypoint level $\phi$ and group shapley $\varphi$ value are normalized to ensure its sum is equal to 1.

\subsection{Group-based Keypoint Removal}
The result of group shapley value indicates that each keypoint only highly related to a subset of keypoints rather than all keypoints.
Hence, we can conclude that when a keypoint is occluded, deep model can infer invisible keypoint based on the information from related keypoints, \eg, the elbow keypoint can be inferred by shoulder and wrist.
Based on this observation, we design Group-based Keypoint Removal (GKR) data augmentation to enhance model's inferential capability of invisible keypoints.

GKR applies random perturbations on individual keypoint within each group while keeping the other keypoints unchanged during training. The perturbation is implemented through replace a randomly selected rectangular region centered at keypoints with random noise.
The detail process of GKR augmentation as follow, 
\begin{algorithm}[t]
\SetAlgoLined
\SetKwInOut{Input}{Input}
\SetKwInOut{Output}{Output}
\SetKwInput{Initialization}{Initialization}
\caption{GKR augmentation Procedure}\label{algorithm 1}
\Input{ Input image $I$; \\
        Image size $W$ and $H$; \\
        Perturb probability $p$;\\
        Group keypoint $\mathcal{P}=\{\mathcal{G}_1, \mathcal{G}_2,..., \mathcal{G}_c\}$;\\
        Perturb area scale $S=\{s_1,s_2,...,s_n\}$.}
\Output{Erased image $I^{\ast}$.}
\Initialization{$p_1 \leftarrow $ Rand (0, 1).} 
\For {$i = 1$ to c}
{
    $p_i \leftarrow $ Rand (0, 1)\\
    \If{$p_i>p$}
    {
        $j\leftarrow$ Randint (1, $|\mathcal{P}_i|$)\\
        $x, y\leftarrow \mathcal{P}_i[j][0]$, $\mathcal{P}_i[j][1]$\\
        $s\leftarrow S[\mathcal{P}_i]$\\
        $h\leftarrow H\times s, w=W\times s$\\
        $I_p\leftarrow(x-\frac{w}{2},y-\frac{h}{2},x+\frac{w}{2},y+\frac{h}{2})$,\\
        $I(I_P)\leftarrow$Rand(0,255)\\
        $I^{\ast}\leftarrow I(I_P)$
    }
}
\end{algorithm}
\vspace{-4mm}

\begin{table*}
\caption{Performance drop with when perturbing individual keypoint.}
\setlength{\tabcolsep}{2.5 pt}
\small
\begin{tabular}{l|cccccccccccccccccc}
\toprule
$\Delta AP$  &baseline&nose &l-eye &r-eye &l-ear &r-ear &l-shd &r-shd &l-elbow &r-elbow &l-wrist &r-wrist &l-hip &r-hip &l-knee &r-knee &l-foot &r-foot    \\
\midrule
nose    &76.1&20.6& 5.3 &4.1 &1.7 &1.3&0.7&0.9&0.0&0.1&0.1&0.0&0.1&0.2&0.2&0.1&0.1&0.0\\
l-eye   &77.5&9.3&7.1&2.3&2.7&2.3&0.6&0.7&0.1&0.4&0.1&0.1&0.2&0.0&0.1&0.1&0&0.0\\
r-eye   &77.5&10.2&3.3&7.5&1.2&1.6&0.6&0.4&0.4&0.2&0.0&0.1&0.2&0.2&0.1&0.1&0.1&0.1\\
l-ear   &79.4&2.4&2.3&1.8&8.9&1.7&0.5&0.2&0.3&0.6&0.5&0.6&0.1&0.1&0.1&0.2&0.2&0.1\\
r-ear   &78.6&2.4&2.8&2.6&1.9&9.3&0.4&0.6&0.1&0.2&0.1&0.0&0.1&0.0&0.1&0.1&0.0&0.0 \\
l-shoulder &81.5&0.5&0.5&0.3&0.4&0.2&14.0&0.2&3.2&0.4&1.9&0.3&0.5&0.2&0.1&0.0&0.1&0.0 \\
r-shoulder &81.6&0.8&0.5&0.4&0.2&0.2&3.2&12.5&0.5&2.5&0.3&1.9&0.2&0.3&0.2&0.3&0.0&0.2 \\
l-elbow &76.4&0.2&0.3&0.0&0.2&0.1&4.8&0.8&15.1&0.5&6.0&0.7&0.9&0.5&0.4&0.0&0.0&0.1 \\
r-elbow &75.7&0.1&0.1&0.0&0.2&0.0&0.9&4.9&0.5&14.5&0.6&6.3&0.2&0.7&0.0&0.2&0.2&0.0 \\
l-wrist &73.2&0.2&0.3&0.1&0.0&0.1&2.9&0.8&9.4&1.0&27.3&1.9&3.0&1.2&0.6&0.1&0.1&0.0 \\
r-wrist &73.1&0.2&0.2&0.0&0.1&0.1&0.6&3.0&1.2&8.2&0.9&27.4&1.0&3.1&0.6&0.0&0.1&0.1 \\
l-hip   &76.0&0.6&0.3&0.3&0.2&0.1&1.6&1.1&0.8&0.6&0.6&0.7&4.3&1.3&2.2&0.5&1.4&0.4 \\
r-hip   &76.0&0.2&0.2&0.1&0.1&0.0&1.2&1.4&0.4&0.6&0.2&0.6&1.4&4.5&0.6&1.9&0.2&1.2 \\
l-knee  &78.1&0.1&0.1&0.0&0.0&0.1&0.2&0.1&0.0&0.0&0.3&0.0&2.3&0.3&8.9&0.8&1.5&0.2 \\
r-knee  &78.5&0&0.2&0&0.0&0.2&0.1&0.3&0.0&0.1&0.3&0.0&0.8&2.7&1.6&10.2&0.3&2.8 \\
l-foot  &79.2&0.1&0.0&0.0&0.0&0.1&0.2&0.1&0.3&0.2&0.3&0.0&1.6&0.5&2.1&0.7&10.6&0.9 \\
r-foot  &79.9&0&0.0&0.1&0.1&0.1&0.4&0.2&0.0&0.0&0.0&0.2&0.3&1.9&0.8&2.1&1.0&10.9 \\
\bottomrule
\end{tabular}
\label{table:ap_drop}
\end{table*}

\section{Experiment}
\subsection{Dataset}
We explore XPose on widely used multi-person pose estimation datasets MSCOCO~\cite{lin2014microsoft}. MSCOCO dataset contains about 80k images and over 200k person instances with 17 annotated body keypoints. COCO dataset is divided into train, val, and test-dev sets with 57k, 5k and 20k images. The performance is evaluated by Average Precision (AP).
Our models are trained on COCO train2017 with about 57k images. All ablation study are conducted on val2017 set.

\subsection{Implementation Details}~\label{sec:implement}
\textbf{Training:} We select 3 typical method to implement XPose, include top-down method HRNet~\cite{sun2019deep}, bottom-up method HigherHRNet~\cite{cheng2020higherhrnet} and regression method PolarPose~\cite{li2023polarpose}. 

For training, We employed the data augmentation mentioned in corresponding paper~\cite{sun2019deep,cheng2020higherhrnet,li2023polarpose}, including random rotation, random scale, random translation and random flipping. 
The input images are resized to $256\times 192$ for HRNet~\cite{sun2019deep} and $512\times 512$ for HigherHRNet~\cite{cheng2020higherhrnet} and PolarPose~\cite{li2023polarpose}. For perturb scale $s$ in GKR data augmentation, we set $s=0.05$ for head keypoints and $s=0.15$ for other keypoints.

\textbf{Test:} Following original paper~\cite{sun2019deep,cheng2020higherhrnet,li2023polarpose}, flip testing is used for all experiments. All reported performance is obtained with single model without ensembling.

\begin{table*}[t]
\begin{minipage}[c]{0.48\textwidth}
\caption{Normalized shapely value of group.}
\setlength{\tabcolsep}{8 pt}
\small
\begin{tabular}{l|ccccc}
\toprule
value(\%)  &head &l-arm &r-arm &l-leg &r-leg    \\
\midrule
head     &96.2&1.4 &1.9 &0.4  &0.1 \\
l-arm    &6.8 &83.1&5.8 &2.7  &1.7 \\
r-arm    &6.0 &6.6 &83.5&1.2  &2.7 \\
l-leg    &4.0 &6.3 &4.1 &69.5 &16.1\\
r-leg    &4.2 &5.3 &5.9 &18.2 &66.4\\
\bottomrule
\end{tabular}
\label{table:shapley1}
\end{minipage}
\hspace{5mm}
\begin{minipage}[c]{0.48\textwidth}
\caption{Normalized shapely value of head keypoint.}
\setlength{\tabcolsep}{8 pt}
\small
\begin{tabular}{l|ccccc}
\toprule
value(\%)  &nose &l-eye &r-eye &l-ear &r-ear    \\
\midrule
nose     &52.4 &13.3 &26.7 &4.4 &3.4 \\
l-eye    &41.6 &31.8 &8.8  &13.0&4.9 \\
r-eye    &44.8 &12.4 &30.3 &4.4 &8.1  \\
l-ear    &10.7 &10.8 &3.8  &65.6&9.1  \\
r-ear    &15.0 &3.6 &12.3 &3.4 &65.7  \\
\bottomrule
\end{tabular}
\label{table:shapley2}
\end{minipage}

\vspace{5mm}
\begin{minipage}[c]{0.23\textwidth}
\caption{Left arm shapely value.}
\setlength{\tabcolsep}{1.2 pt}
\small
\begin{tabular}{l|ccc}
\toprule
value  &l-shd &l-elbow &l-wrist  \\
\midrule
l-shoulder  &74.2 &18.6 &7.2  \\
l-elbow    &25.3 &50.3 &24.4  \\
r-wrist    &8.7  &23.7 &67.7  \\
\bottomrule
\end{tabular}
\label{table:shapley3}
\end{minipage}
\hspace{3mm}
\begin{minipage}[c]{0.23\textwidth}
\caption{right arm shapely value.}
\setlength{\tabcolsep}{1.2 pt}
\small
\begin{tabular}{l|ccc}
\toprule
value  &r-shd &r-elbow &r-wrist  \\
\midrule
r-shoulder  &73.5 &17.9 &8.6  \\
r-elbow    &24.6 &49.6 &25.8  \\
r-wrist    &9.0  &22.5 &68.5  \\
\bottomrule
\end{tabular}
\label{table:shapley4}
\end{minipage}
\hspace{3mm}
\begin{minipage}[c]{0.23\textwidth}
\caption{Left leg shapely value.}
\setlength{\tabcolsep}{1.2 pt}
\small
\begin{tabular}{l|ccc}
\toprule
value  &l-hip &l-knee &l-ankle  \\
\midrule
l-hip   &68.8 &23.8 &7.3  \\
l-knee  &14.9 &65.0 &20.1  \\
l-ankle &6.7  &25.5 &67.9  \\
\bottomrule
\end{tabular}
\label{table:shapley5}
\end{minipage}
\hspace{3mm}
\begin{minipage}[c]{0.23\textwidth}
\caption{Right leg shapely value.}
\setlength{\tabcolsep}{1.2 pt}
\small
\begin{tabular}{l|ccc}
\toprule
value  &r-hip &r-knee &r-ankle  \\
\midrule
r-hip   &72.4 &20.3 &7.3  \\
r-knee  &14.3 &66.1 &19.6  \\
r-ankle &6.7  &23.4 &69.1  \\
\bottomrule
\end{tabular}
\label{table:shapley6}
\end{minipage}
\end{table*}

\begin{figure}
\centering
\includegraphics[width=0.9\linewidth]{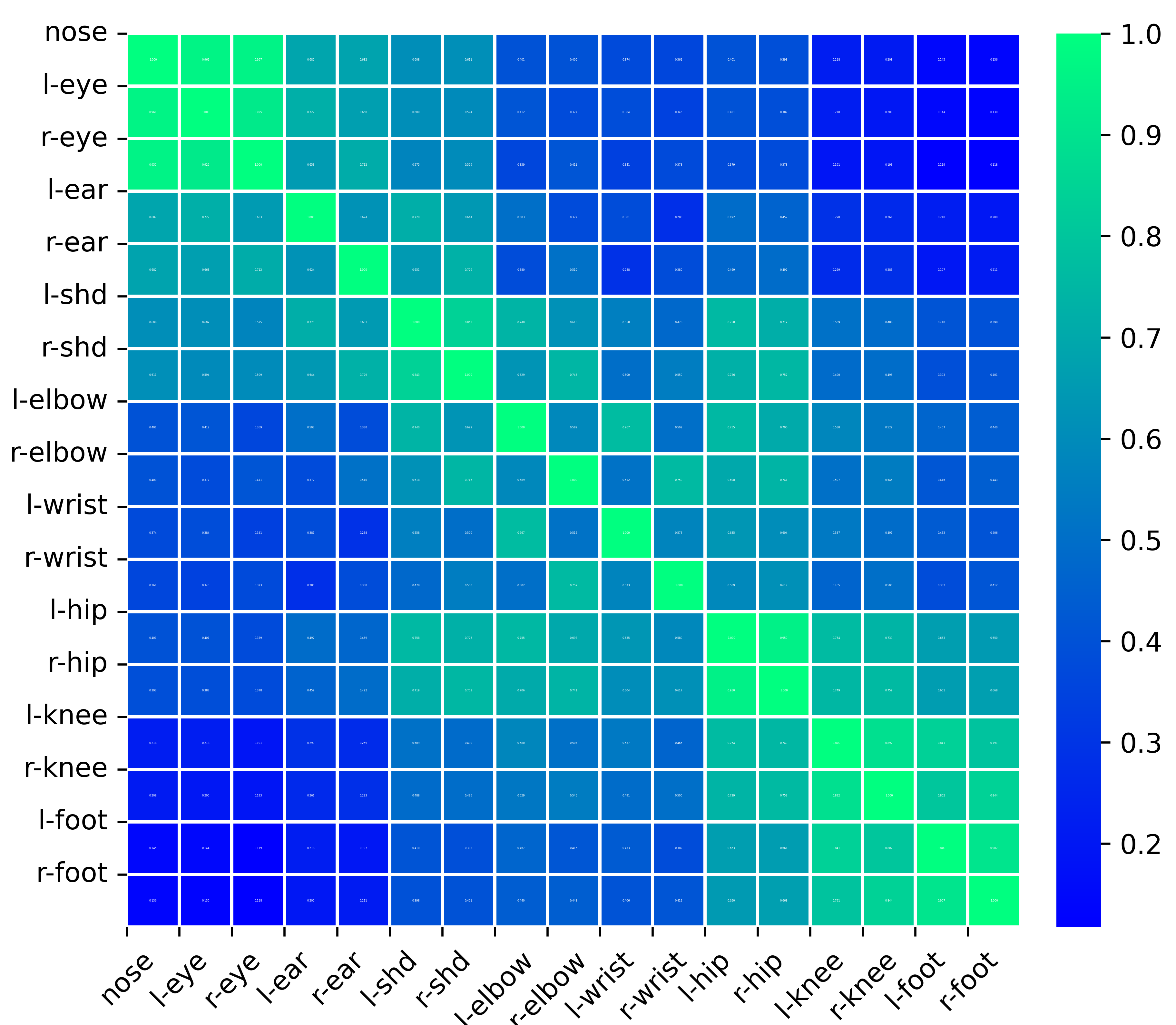}\\
\caption{The correlation of predicted keypoint confidence on whole dataset. The high correlation value denotes the 2 keypoint concurrently receive high or low confidence scores.}
\label{fig:confidence}
\end{figure}

\subsection{Experimental result and analysis}~\label{sec:experiment}
This section exhibites the experimental result and provide detail explaination and analysis.

\textbf{The Interdependencies of each keypoint:} This part investigates the interdependencies of each keypoint pair. Experimental results are summarized in Table~\ref{table:ap_drop}. Each line in Table~\ref{table:ap_drop} shows the performance of individual keypoint, and each column indicates perturbing corresponding keypoint.

As shown Table~\ref{table:ap_drop} that, the largest peformance drop of each keypoint comes from removing itself, \eg, nose keypoint suffer 20.6\% performance drop when remove itself. 
While perturbing a single keypoint does not decrease its performance to near zero, this shows deep model could infer invisible keypoints based on contextual cues from other visible keypoints. Table~\ref{table:ap_drop} also shows that, besides itself, keypoints are most influenced by physiologically connected points, \eg, the elbow keypoint is mostly influenced by shoulder and wrist.

\textbf{Group Shapley Value:} 
Tthis section investigates the GSV.
Table~\ref{table:shapley1} shows the group-level shapley value and Table~\ref{table:shapley2}-~\ref{table:shapley6} summarizes the intra-group shapley inside each group. 
To better observe the contribution of keypoints, we normalize shapley values to make their sum equal to 1.

It can be observed from Table~\ref{table:shapley1} that, there is little correlation between groups, \eg, left arm information only contribute 1.4\% in head keypoints prediction. It can also observed from table, there is non-negligible correlation  between the keypoints of the left and right leg.
The reason is because that, the keypoints in left and right leg are usually overlap by each other, the removal operation on individual keypoint may also remove other keypoints, hence make left and right legs have relatively high correlations.

Table~\ref{table:shapley2} shows the normalize the shapley value of head keypoints. It can be observed from table that, nose and eyes have high correlation, the reason is that the eyes and nose keypoints are close in proximity, making it arbitrary to predict one keypoint based on the other.
While the ear keypoints are relatively far from other keypoints, and predicting their position based on other keypoints can only be done vaguely.

Table~\ref{table:shapley3} and Table~\ref{table:shapley4} summarised the shapley value on left and right arms keypoints. Compared with head keypoints, the arm keypoints have a closer correlation with itself, \eg, left shoulder keypoint contribute 74.2\% to left shoulder accuracy.
It's interesting to observe that, the correlation between keypoints is consistent with the anatomical structure of the human body, \eg, the shoulder keypoint have high correlation with elbow keypoint, while low correlation with wrist keypoint, the elbow keypoint has high correlation with both shoulder and wrist keypoints. The reason is that the elbow keypoint is directly connected to both shoulder and wrist keypoints, allowing the inference of these two keypoints based on the human body structure, whereas the shoulder and elbow are not directly connected, making it difficult to infer.

\begin{figure*}
\centering
\includegraphics[width=0.98\linewidth]{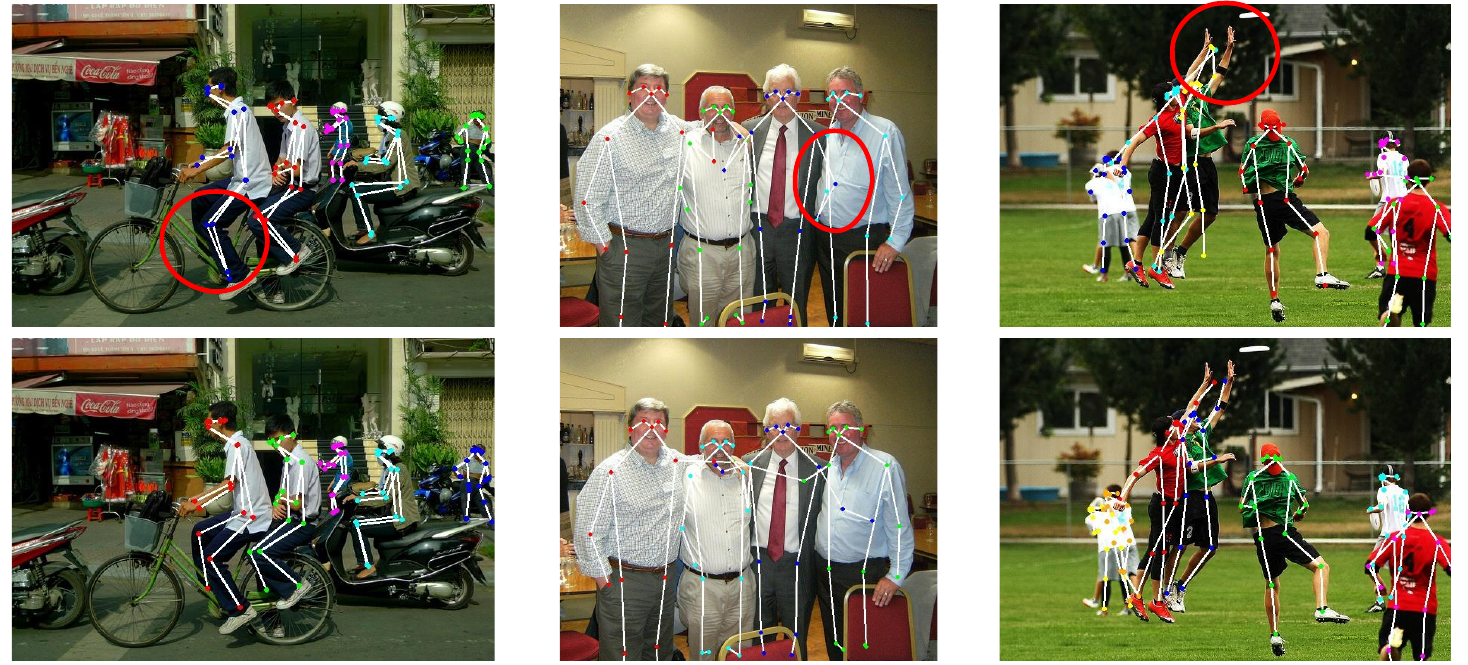}\\
\caption{Illustration of pose estimation results achieved by HRNet (1st line) and HRNet+GKR (2nd line), keypoins of different persons are annotated with different colors.}
\label{fig:visualization}
\end{figure*}

Table~\ref{table:shapley5} and Table~\ref{table:shapley6} summarised the shapley value on left and right legs keypoints. Similar with arms keypoints, the leg keypoints is also consistent with the anatomical structure of human body. The contribution of knee to hip keypoint is relatively low, mainly because the hip keypoint area is larger, leading to significant errors when inferring based on the knee.


Based on inter and intra-group shapley value, we elucidate the comprehensive contributions of each keypoint to final prediction. Specifically, within the same group, we leverage fine-grained intra-group Shapley values to elucidate contributions of individual keypoint. Meanwhile, for keypoints belonging to different groups, we utilize the inter-group Shapley values associated with their respective groups. The overall contribution is 
We present the overall Shapley value and Shapley value based on individual image in appendix.

\textbf{Correlation of Confidence score:} This part exhibits the correlation of predicted keypoint confidence score on overall dataset, the result is illustrated in Fig.~\ref{fig:confidence}. 
The correlation between keypoints is computed based on all instances on COCO training set. A high correlation indicates that these two keypoints concurrently receive either high or low confidence scores.
As shown in Fig.~\ref{fig:confidence}, the correlation of confidence score exhibits similar conclusion with grouping result in Table~\ref{table:group}, \eg, the confidence of keypoints in same group is relatively high.  
The reason is that deep model can infer a keypoint based on its related keypoints, resulting in simultaneous high/low confidence scores for multiple keypoints. 

\begin{table}
\caption{The effectiveness of GKR on COCO dataset. }
\setlength{\tabcolsep}{8 pt}
\small
\begin{tabular}{l|c|cccc}
\toprule
\multirow{2}{*}{Method} &\multirow{2}{*}{data aug} &\multicolumn{4}{c}{occlusion ratio}  \\\cline{3-6}
&&0 &0.25 &0.5 &0.75\\
\midrule
        &None &74.5 &68.2 &54.6 &14.9\\
HRNet   &RE   &74.9 &68.5 &55.0 &15.3\\
        &GKR  &75.5 &68.9 &55.4 &15.5\\\midrule
       &None  &66.4 &60.0 &43.8 &9.2\\
HrHRNet&RE    &66.7 &60.5 &44.5 &9.6\\
       &GKR   &67.0 &61.0 &44.7 &9.9\\\midrule
         &None&68.5 &62.1 &47.3 &10.3\\
PolarPose&RE  &68.9 &62.7 &47.7 &10.6\\
         &GKR &69.3 &63.3 &47.9 &11.0\\\bottomrule
\end{tabular}
\label{table:dataaug}
\end{table}


\subsection{Evaluation of GKR data augmentation}
This part verifies the effectiveness of GKR. We divide COCO dataset into subsets with different occlusion ratio. The occlusion ratio is defined as the ratio of invisible keypoint to all keypoint in an image. 
We implement 3 typical methods to verify GKR, include HRNet~\cite{sun2019deep}, HigherHRNet~\cite{cheng2020higherhrnet} and PolarPose~\cite{li2023polarpose}. 
The experimental results are summarized in Table~\ref{table:dataaug}. In Table~\ref{table:dataaug}, ``None" denotes the baseline method which directly trains deep model without any data augmentation, ``RE" applies random erasing data agumentation~\cite{zhong2020random}, and ``GKR" applies GKR data augmentation for training.

In Table~\ref{table:dataaug}, based on baseline ``None", both RE and GKR promote performance on non-occlusion data, \eg, RE promotes baseline by 0.4\% on HRNet~\cite{sun2019deep}. GKB promotes the perofrmance by 1.0\%, which is more than twice of RE. 
The reason is because that, without the guidance of keypoints, random erasing data augmentation may simultaneously remove multiple related keypoints, which confusing the model training.
In contrast, our approach removes single keypoint from each group, allowing model infer deleted keypoint based on remaining keypoints, hence strength the inferential capability.
We further verify GKR on samples with occlusion. As shown in  Table~\ref{table:dataaug}, 
under the occlusion scenarios, GKR demonstrates superior accuracy.This shows the effectiveness of GKR to enhence the inferential capability for invisible keypoints.

Finally, we illustrate some visualization results in Fig.~\ref{fig:visualization} achieved by HRNet~\cite{sun2019deep} and HRNet+GKR. We emphasize the difference between predictions in the figure.
As shown in the figure, the result achieved by GKR can better predict invisible body keypoints.

\section{Conclusion}
This paper introduces XPose, a framework that integrates XAI into pose estimation task to elucidate the individual contribution of each keypoint to the final predictions. XPose assesses the contribution of each keypoint through a novel group Shapley value, designed to circumvent the complexities associated with the original Shapley value computation. Building on XPose, we also propose a group-based keypoint removal data augmentation method to enhance the model's inferential capability for invisible keypoints. Comprehensive experimental results across three typical methods showcase the effectiveness of the proposed data augmentation approach.




\clearpage
\bibliographystyle{named}
\bibliography{ijcai24}

\begin{thebibliography}{}

\bibitem[\protect\citeauthoryear{Cao \bgroup \em et al.\egroup
  }{2017}]{cao2017realtime}
Zhe Cao, Tomas Simon, Shih-En Wei, and Yaser Sheikh.
\newblock Realtime multi-person 2d pose estimation using part affinity fields.
\newblock In {\em CVPR}, pages 7291--7299, 2017.

\bibitem[\protect\citeauthoryear{Chang \bgroup \em et al.\egroup
  }{2019}]{chang2019explaining}
Chun-Hao Chang, Elliot Creager, Anna Goldenberg, and David Duvenaud.
\newblock Explaining image classifiers by counterfactual generation, 2019.

\bibitem[\protect\citeauthoryear{Cheng \bgroup \em et al.\egroup
  }{2020}]{cheng2020higherhrnet}
Bowen Cheng, Bin Xiao, Jingdong Wang, Honghui Shi, Thomas~S Huang, and Lei
  Zhang.
\newblock Higherhrnet: Scale-aware representation learning for bottom-up human
  pose estimation.
\newblock In {\em CVPR}, pages 5386--5395, 2020.

\bibitem[\protect\citeauthoryear{Covert \bgroup \em et al.\egroup
  }{2021}]{removal_based_XAI_survey}
Ian~C. Covert, Scott Lundberg, and Su-In Lee.
\newblock Explaining by removing: a unified framework for model explanation.
\newblock {\em J. Mach. Learn. Res.}, 22(1), jan 2021.

\bibitem[\protect\citeauthoryear{Devlin \bgroup \em et al.\egroup
  }{2018}]{devlin2018bert}
Jacob Devlin, Ming-Wei Chang, Kenton Lee, and Kristina Toutanova.
\newblock Bert: Pre-training of deep bidirectional transformers for language
  understanding.
\newblock {\em arXiv preprint arXiv:1810.04805}, 2018.

\bibitem[\protect\citeauthoryear{Fang \bgroup \em et al.\egroup
  }{2017}]{fang2017rmpe}
Hao-Shu Fang, Shuqin Xie, Yu-Wing Tai, and Cewu Lu.
\newblock Rmpe: Regional multi-person pose estimation.
\newblock In {\em ICCV}, pages 2334--2343, 2017.

\bibitem[\protect\citeauthoryear{Fong \bgroup \em et al.\egroup
  }{2019}]{DBLP:conf/iccv/FongPV19}
Ruth Fong, Mandela Patrick, and Andrea Vedaldi.
\newblock Understanding deep networks via extremal perturbations and smooth
  masks.
\newblock In {\em 2019 {IEEE/CVF} International Conference on Computer Vision,
  {ICCV} 2019, Seoul, Korea (South), October 27 - November 2, 2019}, pages
  2950--2958. {IEEE}, 2019.

\bibitem[\protect\citeauthoryear{Guyon and
  Elisseeff}{2003}]{10.5555/944919.944968}
Isabelle Guyon and Andr\'{e} Elisseeff.
\newblock An introduction to variable and feature selection.
\newblock {\em J. Mach. Learn. Res.}, 3(null):1157–1182, mar 2003.

\bibitem[\protect\citeauthoryear{He \bgroup \em et al.\egroup
  }{2016}]{he2016deep}
Kaiming He, Xiangyu Zhang, Shaoqing Ren, and Jian Sun.
\newblock Deep residual learning for image recognition.
\newblock In {\em Proceedings of the IEEE conference on computer vision and
  pattern recognition}, pages 770--778, 2016.

\bibitem[\protect\citeauthoryear{He \bgroup \em et al.\egroup
  }{2017}]{he2017mask}
Kaiming He, Georgia Gkioxari, Piotr Doll{\'a}r, and Ross Girshick.
\newblock Mask r-cnn.
\newblock In {\em CVPR}, pages 2961--2969, 2017.

\bibitem[\protect\citeauthoryear{Ignatiev}{2020}]{ijcai2020p726}
Alexey Ignatiev.
\newblock Towards trustable explainable ai.
\newblock In Christian Bessiere, editor, {\em Proceedings of the Twenty-Ninth
  International Joint Conference on Artificial Intelligence, {IJCAI-20}}, pages
  5154--5158. International Joint Conferences on Artificial Intelligence
  Organization, 7 2020.
\newblock Early Career.

\bibitem[\protect\citeauthoryear{Krizhevsky \bgroup \em et al.\egroup
  }{2012}]{krizhevsky2012imagenet}
Alex Krizhevsky, Ilya Sutskever, and Geoffrey~E Hinton.
\newblock Imagenet classification with deep convolutional neural networks.
\newblock {\em Advances in neural information processing systems}, 25, 2012.

\bibitem[\protect\citeauthoryear{Li \bgroup \em et al.\egroup
  }{2023}]{li2023polarpose}
Jianing Li, Yaowei Wang, and Shiliang Zhang.
\newblock Polarpose: Single-stage multi-person pose estimation in polar
  coordinates.
\newblock {\em IEEE Transactions on Image Processing}, 32:1108--1119, 2023.

\bibitem[\protect\citeauthoryear{Lin \bgroup \em et al.\egroup
  }{2014}]{lin2014microsoft}
Tsung-Yi Lin, Michael Maire, Serge Belongie, James Hays, Pietro Perona, Deva
  Ramanan, Piotr Doll{\'a}r, and C~Lawrence Zitnick.
\newblock Microsoft coco: Common objects in context.
\newblock In {\em ECCV}, pages 740--755. Springer, 2014.

\bibitem[\protect\citeauthoryear{Lundberg \bgroup \em et al.\egroup
  }{2020}]{TreeSHAP}
Scott~M. Lundberg, Gabriel Erion, and Hugh et~al. Chen.
\newblock From local explanations to global understanding with explainable ai
  for trees.
\newblock {\em Nature Machine Intelligence}, 2:56--67, 2020.

\bibitem[\protect\citeauthoryear{Newell \bgroup \em et al.\egroup
  }{2017}]{newell2017associative}
Alejandro Newell, Zhiao Huang, and Jia Deng.
\newblock Associative embedding: End-to-end learning for joint detection and
  grouping.
\newblock In {\em NeurIPS}, pages 2277--2287, 2017.

\bibitem[\protect\citeauthoryear{Nie \bgroup \em et al.\egroup
  }{2019}]{nie2019single}
Xuecheng Nie, Jiashi Feng, Jianfeng Zhang, and Shuicheng Yan.
\newblock Single-stage multi-person pose machines.
\newblock In {\em ICCV}, pages 6951--6960, 2019.

\bibitem[\protect\citeauthoryear{Petsiuk \bgroup \em et al.\egroup
  }{2018}]{petsiuk2018rise}
Vitali Petsiuk, Abir Das, and Kate Saenko.
\newblock Rise: Randomized input sampling for explanation of black-box models.
\newblock {\em arXiv preprint arXiv:1806.07421}, 2018.

\bibitem[\protect\citeauthoryear{Qiu \bgroup \em et al.\egroup
  }{2022}]{mypaper}
Luyu Qiu, Yi~Yang, Caleb~Chen Cao, Yueyuan Zheng, Hilary Ngai, Janet Hsiao, and
  Lei Chen.
\newblock Generating perturbation-based explanations with robustness to
  out-of-distribution data.
\newblock In {\em Proceedings of the ACM Web Conference 2022}, WWW '22, page
  3594–3605, New York, NY, USA, 2022. Association for Computing Machinery.

\bibitem[\protect\citeauthoryear{Ren \bgroup \em et al.\egroup
  }{2015}]{ren2015faster}
Shaoqing Ren, Kaiming He, Ross Girshick, and Jian Sun.
\newblock Faster r-cnn: Towards real-time object detection with region proposal
  networks.
\newblock In {\em NeurIPS}, pages 91--99, 2015.

\bibitem[\protect\citeauthoryear{Ribeiro \bgroup \em et al.\egroup
  }{2016}]{lime}
Marco~T{\'{u}}lio Ribeiro, Sameer Singh, and Carlos Guestrin.
\newblock "why should {I} trust you?": Explaining the predictions of any
  classifier.
\newblock In Balaji Krishnapuram, Mohak Shah, Alexander~J. Smola, Charu~C.
  Aggarwal, Dou Shen, and Rajeev Rastogi, editors, {\em Proceedings of the 22nd
  {ACM} {SIGKDD} International Conference on Knowledge Discovery and Data
  Mining, San Francisco, CA, USA, August 13-17, 2016}, pages 1135--1144. {ACM},
  2016.

\bibitem[\protect\citeauthoryear{Sarp \bgroup \em et al.\egroup
  }{2023}]{sarp2023xai}
Salih Sarp, Ferhat~Ozgur Catak, Murat Kuzlu, Umit Cali, Huseyin Kusetogullari,
  Yanxiao Zhao, Gungor Ates, and Ozgur Guler.
\newblock An xai approach for covid-19 detection using transfer learning with
  x-ray images.
\newblock {\em Heliyon}, 9(4):e15137, 2023.

\bibitem[\protect\citeauthoryear{Shapley}{1951}]{shapley1951notes}
Lloyd~S Shapley.
\newblock Notes on the n-person game—ii: The value of an n-person game.
\newblock 1951.

\bibitem[\protect\citeauthoryear{Sun \bgroup \em et al.\egroup
  }{2018}]{sun2018integral}
Xiao Sun, Bin Xiao, Fangyin Wei, Shuang Liang, and Yichen Wei.
\newblock Integral human pose regression.
\newblock In {\em ECCV}, pages 529--545, 2018.

\bibitem[\protect\citeauthoryear{Sun \bgroup \em et al.\egroup
  }{2019}]{sun2019deep}
Ke~Sun, Bin Xiao, Dong Liu, and Jingdong Wang.
\newblock Deep high-resolution representation learning for human pose
  estimation.
\newblock In {\em CVPR}, pages 5693--5703, 2019.

\bibitem[\protect\citeauthoryear{Vaswani and
  others}{2017}]{vaswani2017attention}
Ashish Vaswani et~al.
\newblock Attention is all you need.
\newblock {\em Advances in neural information processing systems}, 30, 2017.

\bibitem[\protect\citeauthoryear{Wei \bgroup \em et al.\egroup
  }{2016}]{wei2016convolutional}
Shih-En Wei, Varun Ramakrishna, Takeo Kanade, and Yaser Sheikh.
\newblock Convolutional pose machines.
\newblock In {\em CVPR}, pages 4724--4732, 2016.

\bibitem[\protect\citeauthoryear{Wei \bgroup \em et al.\egroup
  }{2020}]{wei2020point}
Fangyun Wei, Xiao Sun, Hongyang Li, Jingdong Wang, and Stephen Lin.
\newblock Point-set anchors for object detection, instance segmentation and
  pose estimation.
\newblock {\em arXiv preprint arXiv:2007.02846}, 2020.

\bibitem[\protect\citeauthoryear{Williamson and Feng}{2020}]{SPVIM}
Brian Williamson and Jean Feng.
\newblock Efficient nonparametric statistical inference on population feature
  importance using shapley values.
\newblock In Hal~Daumé III and Aarti Singh, editors, {\em Proceedings of the
  37th International Conference on Machine Learning}, volume 119 of {\em
  Proceedings of Machine Learning Research}, pages 10282--10291. PMLR, 13--18
  Jul 2020.

\bibitem[\protect\citeauthoryear{Winter}{2002}]{shapleyValue}
Eyal Winter.
\newblock Chapter 53 the shapley value.
\newblock volume~3 of {\em Handbook of Game Theory with Economic Applications},
  pages 2025--2054. Elsevier, 2002.

\bibitem[\protect\citeauthoryear{Yoon \bgroup \em et al.\egroup
  }{2019}]{yoon2018invase}
Jinsung Yoon, James Jordon, and Mihaela van~der Schaar.
\newblock {INVASE}: Instance-wise variable selection using neural networks.
\newblock In {\em International Conference on Learning Representations}, 2019.

\bibitem[\protect\citeauthoryear{Zeiler and Fergus}{2014}]{Visualizing2014}
Matthew~D. Zeiler and Rob Fergus.
\newblock Visualizing and understanding convolutional networks.
\newblock In David Fleet, Tomas Pajdla, Bernt Schiele, and Tinne Tuytelaars,
  editors, {\em Computer Vision -- ECCV 2014}, pages 818--833, Cham, 2014.
  Springer International Publishing.

\bibitem[\protect\citeauthoryear{Zhang \bgroup \em et al.\egroup
  }{2020}]{zhang2020distribution}
Feng Zhang, Xiatian Zhu, Hanbin Dai, Mao Ye, and Ce~Zhu.
\newblock Distribution-aware coordinate representation for human pose
  estimation.
\newblock In {\em Proceedings of the IEEE/CVF conference on computer vision and
  pattern recognition}, pages 7093--7102, 2020.

\bibitem[\protect\citeauthoryear{Zhong \bgroup \em et al.\egroup
  }{2020}]{zhong2020random}
Zhun Zhong, Liang Zheng, Guoliang Kang, Shaozi Li, and Yi~Yang.
\newblock Random erasing data augmentation.
\newblock In {\em Proceedings of the AAAI conference on artificial
  intelligence}, volume~34, pages 13001--13008, 2020.

\bibitem[\protect\citeauthoryear{Zhou \bgroup \em et al.\egroup
  }{2015}]{zhou2015object}
Bolei Zhou, Aditya Khosla, Agata Lapedriza, Aude Oliva, and Antonio Torralba.
\newblock Object detectors emerge in deep scene cnns, 2015.

\bibitem[\protect\citeauthoryear{Zhou \bgroup \em et al.\egroup
  }{2019}]{zhou2019objects}
Xingyi Zhou, Dequan Wang, and Philipp Kr{\"a}henb{\"u}hl.
\newblock Objects as points.
\newblock {\em arXiv preprint arXiv:1904.07850}, 2019.

\end{thebibliography}

\end{document}